\documentclass[11pt]{article}

\usepackage[preprint]{acl}

\usepackage{times}
\usepackage{latexsym}

\usepackage[T1]{fontenc}

\usepackage[utf8]{inputenc}

\usepackage{microtype}

\usepackage{inconsolata}

\usepackage{graphicx}
\usepackage{float}
\usepackage{booktabs}
\usepackage{amsmath}
\usepackage{enumitem}

\newcommand{\benchname}{MMGist}

\usepackage[most]{tcolorbox}
\definecolor{badgtcolor}{RGB}{192,57,43}
\definecolor{ambigcolor}{RGB}{211,132,30}
\definecolor{scorercolor}{RGB}{127,83,165}
\definecolor{weakcolor}{RGB}{41,128,185}
\definecolor{satcolor}{RGB}{108,122,137}

\title{MMGist: A Comprehensive Multimodal Benchmark for 2027}

\author{
  Wenzhen Yuan$^{1*\dagger}$ \quad
  Jiacheng Ruan$^{1*}$ \quad
  Wutao Xiong$^{2}$ \quad
  Chengping Zhao$^{2}$ \quad
  Ting Liu$^{1}$ \quad
  Yuzhuo Fu$^{1}$ \\
  \normalfont $^{1}$Shanghai Jiao Tong University \\
  \normalfont $^{2}$Sichuan University
}

\begin{document}
\maketitle
\begingroup
\renewcommand{\thefootnote}{\fnsymbol{footnote}}
\footnotetext[1]{These authors contributed equally.}
\footnotetext[2]{\texttt{winston\_yuan@sjtu.edu.cn}}
\endgroup
\begin{abstract}

We conduct a systematic study of 18 widely used vision-language benchmarks and identify three major issues: \textbf{1)} many items do not rely on visual cues and therefore fail to effectively measure multimodal understanding; \textbf{2)} many items are already close to performance saturation for current LVLMs, which limits their discriminative power; \textbf{3)} a small number of anomalous items affect the reliability of evaluation results. To this end, we propose \textbf{MMGist}, a curated benchmark that covers seven capability dimensions and contains 7,262 items. MMGist is constructed through a three-stage pipeline, which sequentially combines text-ablation filtering, cross-model saturation filtering, and anomaly detection filtering. We conduct extensive experiments on 27 leading LVLMs and compare MMGist with the raw pool of 23,250 items. The results show that MMGist preserves model rankings with high fidelity, with Spearman $\rho = 0.98$, while reducing evaluation items by 69\% and improving cross-model discrimination by 78\%. Further results indicate that Visual Logic remains a systematic weakness of current LVLMs, while knowledge-intensive dimensions such as Expert Knowledge dimensions  remain important factors for distinguishing closed-source models from open-source models. These findings suggest that high-quality evaluation should prioritize visual dependency, discriminative power, and reliability, rather than simply pursuing benchmark scale. The data can be found at \url{https://huggingface.co/datasets/Winston-Yuan/MMGist}.

\end{abstract}

\section{Introduction}

\begin{figure}[t]
    \centering
    \includegraphics[width=\columnwidth]{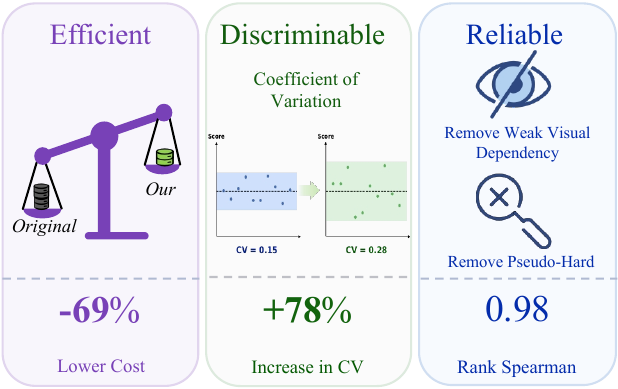}
    \caption{\benchname{} improves multimodal evaluation along three axes: efficiency (69\% fewer items), discrimination (CV $+$78\%), and reliability (removal of low-quality items), while preserving model rankings (Spearman $\rho{=}0.98$).}
    \label{fig:teaser}
\end{figure}

Evaluation of large vision-language models (LVLMs) has shifted from benchmark scarcity to a harder question: how to obtain trustworthy results. Recent benchmarks span mathematical reasoning~\cite{MathVista, MathVision, DynaMath}, document comprehension~\cite{AI2D, OCRBench}, and other capabilities~\cite{MMMU, MMMU-Pro, MME-SCI, LogicVista, HallusionBench, SLAKE, MedXpertQA}, offering an important basis for model comparison~\cite{li2024survey, li2025benchmark}. However, as benchmark suites grow in number and scale, their usefulness increasingly hinges on quality: whether an item truly requires the image~\cite{chen2024we}, still distinguishes model performances~\cite{akhtar2026ai}, and has reliable ground truth and scoring protocols~\cite{MMLU-Redux, northcutt2021pervasive}. When these conditions fail, adding more items increases evaluation cost and can mix non-visual shortcuts, saturated samples, or annotation noise into the final score.

We first conduct an item-level audit of 18 widely used multimodal benchmarks and identify three systematic distortions in existing evaluation signals. The first is weak visual dependence, where models can answer questions from textual cues alone without genuinely understanding image content~\cite{chen2024we}. The second is item saturation, where many items are solved consistently by models with different capability levels and thus contribute little to model discrimination~\cite{akhtar2026ai, tinybenchmarks}. The third is pseudo-hardness, where perceived item difficulty arises from incorrect gold answers, ambiguity, or fragile grading mechanisms~\cite{northcutt2021pervasive}. These issues affect more than evaluation reliability. Running a single evaluation of Qwen3.5-27B on these 18 benchmarks requires 775 GPU hours, while our audit shows that at least 68.8\% of the items are affected by at least one distortion type. In other words, the current evaluation pipeline consumes substantial computational resources while a considerable fraction of the resulting evaluation signals remains distorted.

These issues suggest that reliable multimodal evaluation requires item-level quality control, not merely larger question pools or higher overall difficulty. Following this idea, we propose \textbf{MMGist}, a curated multimodal benchmark built from 18 source benchmarks, containing 7,262 items and covering seven capability dimensions. MMGist maps each type of evaluation distortion to a specific filtering step: text ablation removes items that do not require visual information, cross-model saturation filtering targets items that poorly distinguish current models, and anomaly detection with human review filters out pseudo-difficult items. In other words, MMGist does not simply evaluate fewer items, but concentrates the evaluation budget on items that require image understanding, better distinguish models, and are more reliable.

We comprehensively evaluate 27 leading LVLMs on MMGist. Using only 31.2\% of the raw question pool (totaling 23,250 questions), MMGist preserves highly consistent model-performance rankings, with Spearman $\rho = 0.98$. After filtering, the models' average score drops by 20.3\%, indicating that weakly vision-dependent, saturated, and pseudo-hard questions in the raw pool indeed obscure true model performance. As illustrated in Figure~\ref{fig:teaser}, MMGist reduces evaluation cost by approximately 69\% while improving cross-model discriminability by 78\%, showing that the retained questions more directly characterize model differences. Finally, even the best-performing Gemini-3.1-Pro reaches only 66.8\% Macro Avg, demonstrating that MMGist remains sufficiently challenging while being more efficient.

Our main contributions are threefold: \textbf{1)} we conduct a systematic audit of 23,250 items from 18 widely used vision-language benchmarks and identify three pervasive issues: weak visual dependency, item saturation, and pseudo-hard questions; \textbf{2)} we propose a reusable item-level quality-control pipeline combining text-only ablation filtering, saturation filtering, rule-based anomaly recall, multi-model adjudication, and human expert review; \textbf{3)} we construct MMGist, a curated benchmark with 7,262 items spanning seven capability dimensions. Experiments on 27 LVLMs show that MMGist enables cost-efficient evaluation, ranking consistency with the raw pool, and more discriminative performance comparison.

\section{Related Work}

\subsection{Evaluation Benchmarks for LVLMs}

The rapid advancement of large vision-language models (LVLMs) has produced a broad ecosystem of multimodal evaluation benchmarks~\cite{li2024survey, li2025benchmark}.
Comprehensive benchmarks such as MMBench, MMMU, and MMMU-Pro offer unified testbeds for general perception, reasoning, and multidisciplinary knowledge~\cite{mmbench, MMMU, MMMU-Pro}, while newer benchmarks target specialized skills such as mathematical and scientific reasoning~\cite{MathVista, MathVision, DynaMath, AI2D, MME-SCI}, spatial understanding~\cite{EmbSpatialBench, CountBench}, visual logic~\cite{LogicVista, zerobench}, visual perception and hallucination detection~\cite{blink, HallusionBench, realworldqa}, document and OCR understanding~\cite{OCRBench}, and medical diagnosis~\cite{SLAKE, MedXpertQA}.
These benchmarks underpin LVLM evaluation, yet they often assume that each item is visually necessary, discriminative, and reliably annotated. Our work starts from this assumption and asks whether item-level quality in existing benchmarks is sufficient to support reliable multimodal evaluation.

\begin{figure*}[t]
    \centering
    \includegraphics[width=\textwidth]{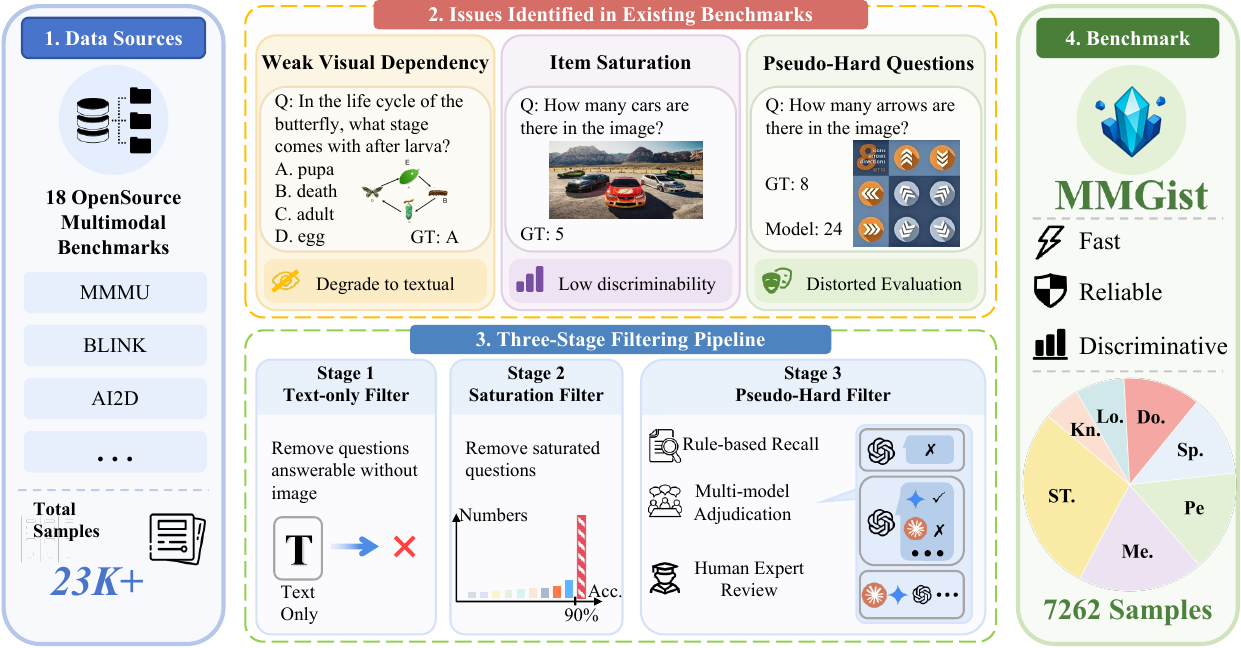}
    \caption{Overview of \benchname{}. We collect 23K+ items from 18 source benchmarks (left), identify three quality issues in existing evaluation: weak visual dependency, item saturation, and pseudo-hard questions (top), apply a three-stage filtering pipeline to remove affected items (bottom), and produce a curated benchmark of 7,262 items spanning seven capability dimensions (right). Capability abbreviations: ST.=STEM Reasoning, Kn.=Expert Knowledge, Lo.=Visual Logic, Do.=Diagram \& OCR, Sp.=Spatial Understanding, Pe.=Visual Perception, Me.=Medical.}
    \label{fig:pipeline}
\end{figure*}

\subsection{Benchmark Quality and Reliability}

Recent work has identified several benchmark-quality risks, but usually studies them in isolation. MMStar and other studies~\citep{chen2024we, brown2025benchmark, xu2024benchmark, deng2024investigating} show that many multimodal items can be answered without images, and related shortcut and contamination studies show that question cues, world knowledge, or training-data overlap may inflate scores. Other work highlights saturation, where already-solved items add cost but little discrimination~\cite{akhtar2026ai, MME-SCI}, and annotation or scoring errors, where even small label-noise rates can change model rankings~\cite{MMLU-Redux, northcutt2021pervasive}.
Harder or dynamic benchmarks, meta-evaluation frameworks, and IRT-based methods improve evaluation, but mainly by adding new items, analyzing individual benchmarks, or targeting one quality dimension~\cite{phan2025humanity, white2024livebench, DyVal, BetterBench, tinybenchmarks, sedoc2020item}.
In contrast, MMGist targets existing multimodal benchmarks with a unified item-level quality-control pipeline that combines text-ablation filtering, cross-model saturation filtering, multi-model adjudication, and expert review to address visual dependency, saturation, and anomalous items jointly.

\section{Problems in Existing Vision-Language Benchmarks}
\label{sec:problems}

\subsection{Weak Visual Dependency}
\label{sec:weak-visual}

Vision-language benchmarks evaluate the multimodal capabilities of LVLMs, assuming that correctly answering each item requires understanding the visual content. However, we find that many items can be answered without accessing the image. We trace this weak visual dependency to two primary sources: (1)~answer leakage from the question text, where phrasing or options make only one answer semantically plausible regardless of the image; and (2)~world knowledge embedded in LLMs, where the question targets factual or commonsense knowledge requiring no image information~\cite{chen2024we}. Both pathways enable correct answers through language-only reasoning, undermining the validity of multimodal evaluation.

To quantify this issue, we use five models from different providers (Qwen3.6-35B-A3B, GPT-5-Mini, Gemini-3.1-Flash-Lite, Doubao-Seed-2.0-Mini, and Claude-Haiku-4.5)~\cite{yang2025qwen3, singh2025openai, comanici2025gemini, guo2025seed1, anthropic2025claude4} as text-only inspectors. Each inspector receives only the image-free question text, and we sample eight responses per item to compute the per-item text-only accuracy. The problem is pervasive: AI2D~\cite{AI2D} exhibits the highest text-only accuracy at 66.7\%, indicating that its items are largely solvable without the diagram. Even expert-knowledge benchmarks, including MMMU~\cite{MMMU} (51.2\%), 
show substantial text-only solvability. This indicates that widely used benchmarks often do not require truly visual understanding.

\subsection{Item Saturation}
\label{sec:saturation}

Beyond visual dependency, many benchmark items are now trivially solvable, with negligible discriminative value among current models. We define an item as \emph{saturated} when models spanning a wide capability range consistently solve it. To assess saturation, we evaluate all 23,250 items with 12 LVLMs from five providers (Anthropic, OpenAI, Google, Alibaba, ByteDance), covering different model tiers.

The results reveal that nearly half the items have lost discriminative power. Across all 18 benchmarks, 49.7\% of items exceed 90\% average accuracy. These items compress the score range, obscuring capability differences and inflating cost.

\subsection{Pseudo-hard Questions}
\label{sec:pseudo-hard}

Not all items that models answer incorrectly are truly challenging. Some failures arise from item-level defects rather than limited model capability. Label errors are well documented in NLP benchmarks: Northcutt et al.~\cite{northcutt2021pervasive} reported an average label-error rate of 3.3\% across ten major benchmarks, enough to alter model rankings. We observed similar signals in vision-language benchmarks: among 23,250 items, 695 received no correct answer from all 12 models, and 27.8\% were manually verified as item-level defects.

We refer to such items as \emph{pseudo-hard} and group them into three types. \emph{Ground-truth errors} occur when the annotated answer is incorrect, penalizing models that reason correctly. \emph{Ambiguous items} involve underspecified questions, non-unique answers, or insufficient evidence, making any answer debatable. \emph{Scorer risks} arise when the scoring protocol cannot reliably match model outputs to the ground truth because of synonymous expressions, mathematical equivalence, rounding discrepancies, or response format mismatches. These pseudo-hard questions distort evaluation by penalizing models for correct reasoning and inflating the apparent difficulty of benchmarks. Together with weak visual dependency and item saturation, these three issues motivate a systematic quality control pipeline.

\section{Constructing \benchname{}}
\label{sec:construction}

Using the three quality issues identified in \S\ref{sec:problems}, we construct \benchname{} from 18 source benchmarks via a multi-stage pipeline (Figure~\ref{fig:pipeline}). The pipeline applies three successive filters, each targeting one issue: text-only ablation filtering for visual necessity (\S\ref{sec:text-only-filter}), saturation filtering to preserve discriminative value (\S\ref{sec:saturation-filter}), and anomaly detection with review to enforce item integrity (\S\ref{sec:anomaly-review}).

\subsection{Text-only Ablation Filtering}
\label{sec:text-only-filter}

Using the text-only inspection setup from \S\ref{sec:weak-visual}, we compute average accuracy over the five models per item and remove items above type-dependent thresholds: 80\% for binary (yes/no) items and 50\% for others. This stage removes 9,465 items, 40.7\% of the original pool.

\subsection{Saturation Filtering}
\label{sec:saturation-filter}

To remove saturated items, we apply the 12 models in \S\ref{sec:saturation} to the 13,785 samples remaining from the previous step. Each model samples each item eight times under the image-conditioned setting, and we compute the average accuracy across models. Items with an average accuracy above 90\% are classified as saturated and removed \footnote{This threshold is intentionally conservative: an item must be consistently solved by almost all models, from lightweight to flagship ones, before being discarded, ensuring that truly challenging items remain even if one or two strong models solve them.}. This stage leaves 8,831 items; CountBench loses 74.1\% of its remaining items, while benchmarks targeting harder reasoning tasks, such as MME-SCI (1.0\%) and ZeroBench (0.0\%), are minimally affected.

\subsection{Anomaly Detection and Review}
\label{sec:anomaly-review}

The final stage targets pseudo-hard questions, whose difficulty stems from ground-truth errors, ambiguity, or scorer risks rather than from genuine visual-language challenges. We adopt a three-step approach: rule-based recall, multi-model adjudication, and human review.

\paragraph{Rule-based recall.}
We designed five heuristic rules to flag suspicious items (Table~\ref{tab:rules}). Each targets a specific failure mode: items scored zero by all 12 models may have ground-truth errors; items whose majority vote consistently disagrees with the labeled answer may contain annotation errors; high inter-model disagreement or near-tied top-vote answers suggest ambiguity; and abnormally low answer-extraction rates indicate scoring-compatibility issues. Applying these rules to 8,831 saturation-filtered samples yielded 5,477 candidates for further review.

\begin{table}[t]
    \centering
    \small
    \begin{tabular}{lp{4.2cm}}
    \toprule
    Rule & Condition \\
    \midrule
    All-fail & Accuracy = 0 across all models \\
    Stable non-GT & Majority vote $\neq$ GT, ratio $>$ 0.5 \\
    Confusion & Top-1 / Top-2 vote ratio $<$ 1.5 \\
    High disagreement & Majority rate $<$ 0.5 or unique answers $\geq$ 10 \\
    Scorer anomaly & Answer extraction rate $<$ 50\% \\
    \bottomrule
    \end{tabular}
    \caption{Risk-assessment rules for anomaly recall. Each rule targets a specific failure mode identified from multi-model sampling metadata.}
    \label{tab:rules}
\end{table}

\paragraph{Multi-model adjudication.}
We employ five LVLMs from different providers (Claude-Sonnet-4.6, Gemini-3.1-Pro, GPT-5, Qwen3.6-Plus, and Doubao-Seed-2.0-Pro)~\cite{anthropic2025claude4, comanici2025gemini, singh2025openai, yang2025qwen3, guo2025seed1} as independent judges in a two-round adjudication process~\cite{zheng2023judging}. In the first round, each judge independently reviews every candidate item, receiving the image, question text, annotated ground truth, and model responses from the sampling stage, and assigns one of four labels: \emph{GT error}, \emph{ambiguous question}, \emph{scorer risk}, or \emph{no issue}. In the second round, each judge receives the aggregated first-round judgments from all five models and re-evaluates the item using the collective evidence. The final decision is reached by majority voting over the second-round labels: an item is confirmed as problematic when at least three judges agree on the non-trivial label.

\paragraph{Human expert review.}
All items deemed problematic by multi-model adjudication are further reviewed by human experts, who verify whether each flagged item is flawed or represents a genuinely difficult challenge. After human review, 1,569 items are confirmed and removed from the benchmark.

\subsection{Final Benchmark}
\label{sec:final-bench}

\begin{figure}[t]
    \centering
    \includegraphics[width=\columnwidth]{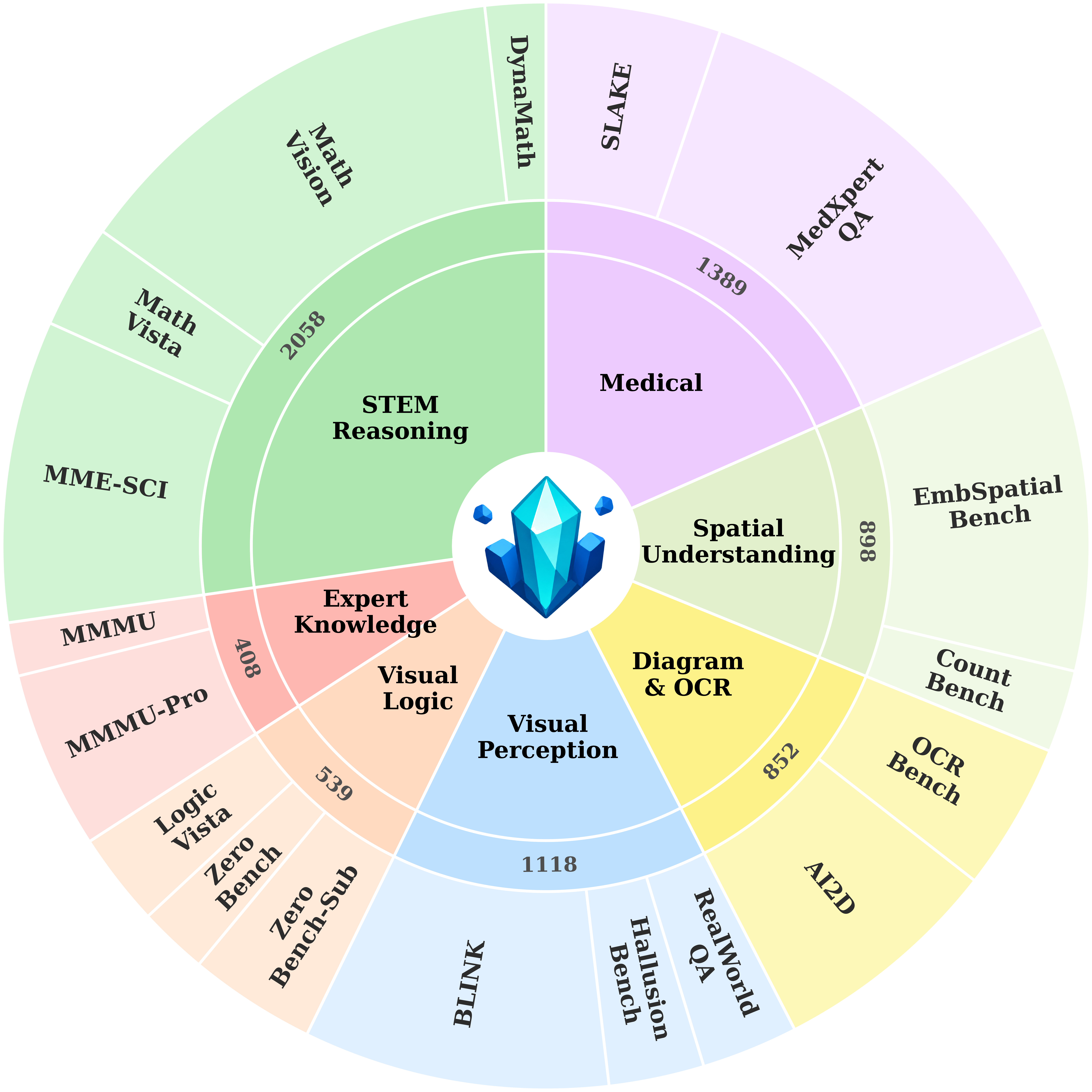}
    \caption{Composition of \benchname{} by capability dimension (inner ring) and source benchmark (outer ring). The 7,262 items span seven categories.}
    \label{fig:capability-pie}
\end{figure}

After all three filtering stages, \benchname{} retains 7,262 items from 18 source benchmarks, representing 31.2\% of the original 23,250 items. The retained items span seven broad capability categories (Figure~\ref{fig:capability-pie}): STEM Reasoning (MathVista, MathVision, DynaMath, MME-SCI)~\cite{MathVista, MathVision, DynaMath, MME-SCI}, Expert Knowledge (MMMU, MMMU-Pro)~\cite{MMMU, MMMU-Pro}, Visual Logic (LogicVista, zerobench, ZeroBench-Sub)~\cite{LogicVista, zerobench}, Visual Perception (BLINK, HallusionBench, RealWorldQA)~\cite{blink, HallusionBench, realworldqa}, Diagram \& OCR (AI2D, OCRBench)~\cite{AI2D, OCRBench}, Spatial Understanding (EmbSpatialBench, CountBench)~\cite{EmbSpatialBench, CountBench}, and Medical (SLAKE, MedXpertQA)~\cite{SLAKE, MedXpertQA}. Retention rates vary substantially across benchmarks, from 96.0\% for ZeroBench, whose items are inherently visually grounded and challenging, to 15.7\% for MMMU, reflecting genuine differences in item quality across existing evaluation suites.

\section{Experiments}
\label{sec:experiments}

\begin{table*}[t]
    \centering
    \small
    \setlength{\tabcolsep}{4.5pt}
    \renewcommand{\arraystretch}{0.88}
    \begin{tabular}{lrrrrrrrrr}
    \toprule
    Model & STEM & Know & Logic & Percep & Doc & Spatial & Med & Macro $\uparrow$ & Sample $\uparrow$ \\
    \midrule
    \multicolumn{10}{l}{\emph{Open-source LVLMs $<$ 10B (Small)}} \\
    
    Gemma 4 E2B            & 15.3 & 22.4 & 11.7 & 37.7 & 38.5 & 31.6 & 18.4 & 25.1 & 23.4 \\
    Gemma 4 E4B            & 18.0 & 25.4 & 13.6 & 39.7 & 44.3 & 41.2 & 21.4 & 29.1 & 26.7 \\
    Kimi-VL-A3B-Instruct   & 15.8 & 24.0 & 10.5 & 36.8 & 42.6 & 34.3 & 14.2 & 25.5 & 23.1 \\
    Kimi-VL-A3B-Thinking   & 19.4 & 18.3 & 11.4 & 37.7 & 35.4 & 37.1 & 16.4 & 25.1 & 23.5 \\
    Ovis2-8B               & 20.1 & 21.4 & 12.3 & 43.2 & 54.5 & 48.8 & 18.7 & 31.3 & 26.7 \\
    InternVL3.5-4B         & 26.3 & 23.2 & 12.5 & 44.1 & 44.5 & 51.0 & 25.6 & 32.5 & 29.6 \\
    InternVL3.5-8B         & 28.2 & 26.3 & 13.9 & 44.1 & 48.5 & 50.5 & 25.1 & 33.8 & 30.4 \\
    Qwen3.5-4B             & 31.1 & 31.9 & 21.0 & 55.6 & 53.7 & 68.7 & 32.1 & 42.0 & 39.1 \\
    Qwen3.5-9B             & 37.1 & 32.7 & 26.4 & 57.6 & 65.5 & 73.5 & 35.0 & 46.8 & 44.0 \\
    Avg. Performance       & 23.5 & 25.1 & 14.8 & 44.1 & 47.5 & 48.5 & 23.0 & 32.4 & 29.6 \\
    \midrule
    \multicolumn{10}{l}{\emph{Open-source LVLMs $\geq$ 10B (Large)}} \\
    Step3-VL-10B           & 36.3 & 38.5 & 21.6 & 53.3 & 63.1 & 65.8 & 24.6 & 43.3 & 40.1 \\
    Ovis2-16B              & 23.4 & 25.0 & 11.8 & 42.5 & 53.1 & 50.5 & 20.1 & 32.4 & 27.9 \\
    Gemma 4 26B-A4B        & 36.8 & 38.7 & 24.3 & 52.1 & 61.3 & 70.0 & 35.5 & 45.6 & 43.4 \\
    Gemma 4 31B            & 40.4 & 45.0 & 27.7 & 55.8 & 63.8 & 73.7 & 39.0 & 49.3 & 47.6 \\
    InternVL3.5-30B-A3B    & 30.6 & 32.8 & 14.1 & 41.7 & 54.9 & 51.5 & 25.5 & 35.9 & 32.4 \\
    InternVL3.5-38B        & 32.5 & 34.9 & 20.3 & 48.7 & 53.1 & 61.5 & 27.5 & 39.8 & 35.9 \\
    Qwen3.5-27B            & 46.6 & 41.7 & 28.1 & 58.0 & 55.4 & 67.8 & 33.2 & 47.3 & 46.3 \\
    Qwen3.6-35B-A3B        & 54.3 & 53.1 & 31.5 & 61.7 & 65.4 & 77.2 & 43.0 & 55.2 & 55.1 \\
    Avg. Performance       & 37.6 & 38.7 & 22.4 & 51.7 & 58.8 & 64.8 & 31.1 & 43.6 & 41.1 \\
    \midrule
    \multicolumn{10}{l}{\emph{Closed-source LVLMs}} \\
    Qwen3.6-Plus           & \textbf{62.6} & 64.4 & 34.9 & 67.3 & 68.7 & 79.3 & 50.0 & 61.0 & 62.0 \\
    Claude Haiku 4.5       & 28.6 & 31.6 & 16.9 & 38.1 & 33.6 & 45.8 & 21.8 & 30.9 & 29.1 \\
    Claude Sonnet 4.6      & 44.6 & 54.9 & 22.4 & 51.2 & 61.3 & 58.2 & 37.5 & 47.2 & 46.5 \\
    Doubao Seed 2.0 Lite   & 53.1 & 53.8 & 35.1 & 65.4 & 63.2 & 74.8 & 39.6 & 55.0 & 54.1 \\
    Doubao Seed 2.0 Mini   & 48.3 & 54.5 & 30.3 & 63.2 & 63.2 & 72.0 & 35.2 & 52.4 & 50.8 \\
    Doubao Seed 2.0 Pro    & 57.6 & 58.4 & 37.7 & 70.2 & 65.8 & 74.9 & 43.0 & 58.2 & 58.2 \\
    GPT-5-Mini             & 36.8 & 49.2 & 24.4 & 52.7 & 54.4 & 60.5 & 37.2 & 45.0 & 44.2 \\
    GPT-5                  & 46.1 & 60.9 & 25.8 & 58.5 & 58.6 & 63.6 & 46.6 & 51.4 & 52.9 \\
    Gemini 3.1 Flash Lite  & 41.5 & 54.1 & 27.2 & 62.4 & 73.4 & 68.2 & 41.8 & 52.7 & 51.5 \\
    Gemini 3.1 Pro        & 61.2 & \textbf{70.2} & \textbf{40.4} & \textbf{75.4} & \textbf{82.3} & \textbf{79.7} & \textbf{58.0} & \textbf{66.8} & \textbf{68.0} \\
    Avg. Performance       & 48.0 & 55.2 & 29.5 & 60.4 & 62.5 & 67.7 & 41.1 & 52.1 & 51.7 \\
    \bottomrule
    \end{tabular}
    \caption{Performance of 27 LVLMs on MMGist across seven capability dimensions. Macro Avg is the mean over capability groups; Sample Avg is the mean over all 7,262 items. Both yield consistent rankings (Spearman $\rho = 0.98$). Best per column in \textbf{bold}.}
    \label{tab:main-results}
\end{table*}

\subsection{Experimental Setup}
\label{sec:exp-setup}

\paragraph{Models.}
We evaluate 27 LVLMs across model families, scales, and deployment regimes. The closed-source models come from five commercial API providers, including Gemini-3.1-Pro and Flash-Lite~\citep{comanici2025gemini}, GPT-5 and GPT-5-Mini~\citep{singh2025openai}, Claude-Sonnet-4.6 and Claude-Haiku-4.5~\citep{anthropic2025claude4}, Qwen3.6-Plus~\citep{yang2025qwen3}, and Doubao-Seed-2.0-Pro/Mini/Lite~\citep{guo2025seed1}. The open-source models span seven families and range from 2B to 38B parameters: Qwen3.6-35B-A3B, Qwen3.5 (4B, 9B, 27B)~\citep{yang2025qwen3}, Gemma-4 (E2B, E4B, 26B-A4B, 31B)~\citep{gemma4modelcard2026}, InternVL3.5 (4B, 8B, 30B-A3B, 38B)~\citep{wang2025internvl3}, Ovis2 (8B, 16B)~\citep{lu2024ovis}, Kimi-VL-A3B (Instruct, Thinking)~\citep{team2025kimi}, and Step3-VL-10B~\citep{huang2026step3}.

\paragraph{Implementation details.}
To keep scores comparable, we use the same evaluation protocol for all models. Each model is queried $R=8$ times per item with temperature $T=1.0$ and a maximum output length of 16,384 tokens. A unified prompt schema asks models to reason step by step and then place the final answer in a \texttt{\textbackslash boxed\{\}} block, which is parsed by the same rule-based extractor across benchmarks. For closed-source models that support configurable inference effort, we set it to \emph{medium}. Open-source models are served via vLLM~\cite{kwon2023efficient} on 8$\times$H200 GPUs. All scores are reported on a 100-point scale.

\subsection{Main Results}
\label{sec:main-results}

Table~\ref{tab:main-results} shows whether MMGist remains challenging and discriminative after item-level filtering. We focus on three questions: whether current models are close to saturation, which capabilities drive the closed/open gap, and whether aggregate scores hide uneven skill profiles.

\paragraph{1) MMGist remains unsaturated for frontier models.}
MMGist remains far from saturated: the strongest model, Gemini-3.1-Pro, reaches 66.8\% Macro Avg, and only two models exceed 60\%. This difficulty spans all model tiers: closed-source systems average 52.1\%, large open-source models average 43.6\%, and small open-source models average 32.4\%. Because the pipeline has already removed weakly visual, saturated, and pseudo-hard items, remaining errors are more likely to reflect unresolved multimodal skills than label noise or adversarial artifacts.

\paragraph{2) The closed/open gap has shifted from perception to grounded expertise.}

The closed/open gap is largest on tasks that pair visual evidence with domain knowledge, not on basic visual extraction. Large open-source models nearly match closed-source systems on Diagram \& OCR (58.8\% vs.\ 62.5\%) and Spatial Understanding (64.8\% vs.\ 67.7\%), but trail by 16.5 points on Expert Knowledge and 10.0 points on Medical. This suggests that current open-source LVLMs are competitive at extracting visible structure but lag when visual evidence must be integrated with domain knowledge and multi-step reasoning.

\paragraph{3) Visual Logic exposes a system-level reasoning bottleneck.}
Visual Logic is the clearest bottleneck: its cross-model average is only 22.5\%, and even the best model reaches only 40.4\%.
The weakness is not explained by model scale or access pattern: the average rises from 14.8\% for small open-source models to 22.4\% for large open-source models and 29.5\% for closed-source models, but all three groups remain far below their own Macro Avg.
The result is a useful warning: better recognition, OCR, and spatial localization do not automatically lead to better reasoning over visual states, relation, and implicit rules.

\paragraph{4) Strong models often have jagged capability profiles.}
Aggregate scores hide large within-model asymmetries, so capability-level reporting is necessary.
Qwen3.6-35B-A3B achieves 77.2\% on Spatial Understanding but only 31.5\% on Visual Logic, a 45.7-point gap within the same model.
Ovis2-8B reaches 54.5\% on Diagram \& OCR, exceeding several larger models on that dimension, yet obtains only 12.3\% on Visual Logic.
These non-monotonic profiles show that parameter count and overall average score do not guarantee balanced multimodal ability.
The seven-dimensional structure of MMGist separates skills that benefit from transferable visual representations from skills that likely require targeted reasoning data, training objectives, or inference mechanisms.

\subsection{Effect of Filtering on Scores and Rankings}
\label{sec:filtering-effect}

We next test whether curation improves the evaluation signal, not just whether it makes the test set smaller and harder. Table~\ref{tab:rank-shift} compares the same 27 models on the raw pool of 23,250 items and on MMGist with 7,262 retained items. The curated set keeps the main ordering from the raw pool: Spearman rank correlation is $\rho = 0.98$ under the benchmark-level average; 21 models change rank by at most one position, 25 change by at most two, and the top three models remain unchanged. Scores also become more conservative. Average performance drops by 20.3 points, which indicates that weakly visual shortcuts and saturated items in the raw pool inflated scores. The drop does not flatten the leaderboard: the coefficient of variation rises from 0.15 to 0.28 ($+$78\%), and the gap between the top-5 and bottom-5 models widens from 26.9 to 32.3 points. MMGist therefore keeps the original ranking signal while spending more of the evaluation budget on items that separate models.

\begin{table}[t]
    \centering
    \scriptsize
    \setlength{\tabcolsep}{3pt}
    \begin{tabular}{lrrcrrc}
    \toprule
    Model & Raw & R$_\text{b}$ & & Curated & R$_\text{a}$ & $\Delta$R \\
    \midrule
    Gemini 3.1 Pro       & 77.3 &  1 && 65.2 &  1 & 0 \\
    Qwen3.6-Plus         & 74.4 &  2 && 60.1 &  2 & 0 \\
    Doubao S. 2.0 Pro    & 73.6 &  3 && 57.7 &  3 & 0 \\
    Doubao S. 2.0 Lite   & 71.5 &  4 && 54.3 &  4 & 0 \\
    Qwen3.6-35B-A3B      & 69.6 &  6 && 54.1 &  5 & $+$1 \\
    Doubao S. 2.0 Mini   & 69.6 &  8 && 51.3 &  6 & $+$2 \\
    Gemini 3.1 Flash Lite  & 69.6 &  7 && 50.6 &  7 & 0 \\
    GPT-5                & 70.0 &  5 && 49.8 &  8 & $-$3 \\
    Gemma 4 31B          & 67.8 &  9 && 47.5 &  9 & 0 \\
    Qwen3.5-27B          & 67.7 & 10 && 46.7 & 10 & 0 \\
    Claude Sonnet 4.6    & 67.0 & 11 && 45.7 & 11 & 0 \\
    Qwen3.5-9B           & 64.7 & 14 && 45.2 & 12 & $+$2 \\
    Gemma 4 26B-A4B      & 65.6 & 13 && 43.8 & 13 & 0 \\
    GPT-5-Mini           & 66.8 & 12 && 43.4 & 14 & $-$2 \\
    Step3-VL-10B         & 62.7 & 15 && 41.9 & 15 & 0 \\
    Qwen3.5-4B           & 61.3 & 17 && 40.4 & 16 & $+$1 \\
    InternVL3.5-38B      & 61.5 & 16 && 38.4 & 17 & $-$1 \\
    InternVL3.5-30B-A3B  & 56.7 & 20 && 34.4 & 18 & $+$2 \\
    InternVL3.5-8B       & 57.6 & 19 && 32.6 & 19 & 0 \\
    InternVL3.5-4B       & 54.8 & 21 && 31.3 & 20 & $+$1 \\
    Ovis2-16B            & 52.5 & 22 && 30.8 & 21 & $+$1 \\
    Claude Haiku 4.5     & 58.3 & 18 && 30.3 & 22 & $-$4 \\
    Ovis2-8B             & 49.6 & 24 && 29.7 & 23 & $+$1 \\
    Gemma 4 E4B          & 51.9 & 23 && 27.6 & 24 & $-$1 \\
    Kimi-VL-A3B-Think.   & 45.8 & 25 && 24.4 & 25 & 0 \\
    Kimi-VL-A3B-Inst.    & 39.7 & 27 && 24.2 & 26 & $+$1 \\
    Gemma 4 E2B          & 45.5 & 26 && 24.0 & 27 & $-$1 \\
    \bottomrule
    \end{tabular}
    \caption{Model scores and rankings before (raw pool) and after MMGist curation. $\Delta$R denotes rank change (positive = rise). Spearman $\rho = 0.98$.}
    \label{tab:rank-shift}
\end{table}

\subsection{Capability-Level Findings}
\label{sec:capability-findings}

\begin{figure}[t]
    \centering
    \includegraphics[width=\columnwidth]{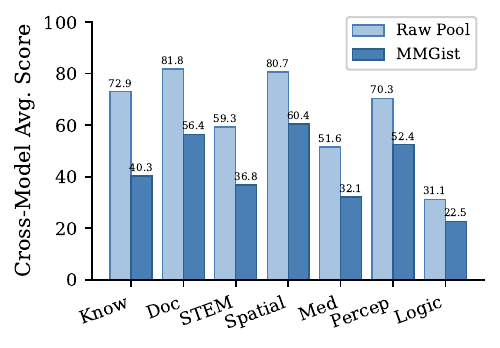}
    \caption{Cross-model average score by capability dimension before and after filtering. Expert Knowledge drops most; Visual Logic drops least.}
    \vspace{-8pt}
    \label{fig:capability-shift}
\end{figure}

Figure~\ref{fig:capability-shift} shows that score inflation in the raw pool varies sharply by capability. Expert Knowledge drops from 72.9\% to 40.3\% ($-$32.6\%), followed by Diagram \& OCR ($-$25.4\%) and STEM Reasoning ($-$22.5\%). These dimensions are more exposed to textual priors, world knowledge, or saturated items. Visual Logic behaves differently: it drops only from 31.1\% to 22.5\% ($-$8.6\%), which suggests that many of its raw-pool items were already visually grounded and unsaturated. After curation, the remaining difficulty sits mostly beyond direct visual extraction. Visual Logic has the lowest cross-model average at 22.5\%, and even the best model reaches only 40.4\%; Medical and STEM Reasoning also remain difficult, averaging 32.1\% and 36.8\%. Spatial Understanding and Diagram \& OCR are higher, at 60.4\% and 56.4\%. This suggests that structured perception, OCR, and spatial localization are comparatively mature, while Visual Logic, Medical, and STEM Reasoning remain the main shared bottlenecks for current LVLMs.

\subsection{Cost Efficiency}
\label{sec:cost-efficiency}

MMGist improves evaluation efficiency through item-level curation, not uniform subsampling. It retains 7,262 of the original 23,250 items (31.2\%), reducing the number of evaluated items by 69\%. It also preserves the raw-pool ranking with Spearman $\rho = 0.98$ and increases cross-model discrimination. In practice, future evaluations can use far fewer model calls while getting rankings that remain stable and easier to separate.

\section{Analysis}
\label{sec:analysis}

\subsection{Error Analysis}
\label{sec:error-analysis}

\begin{figure}[t]
    \centering
    \includegraphics[width=0.9\columnwidth]{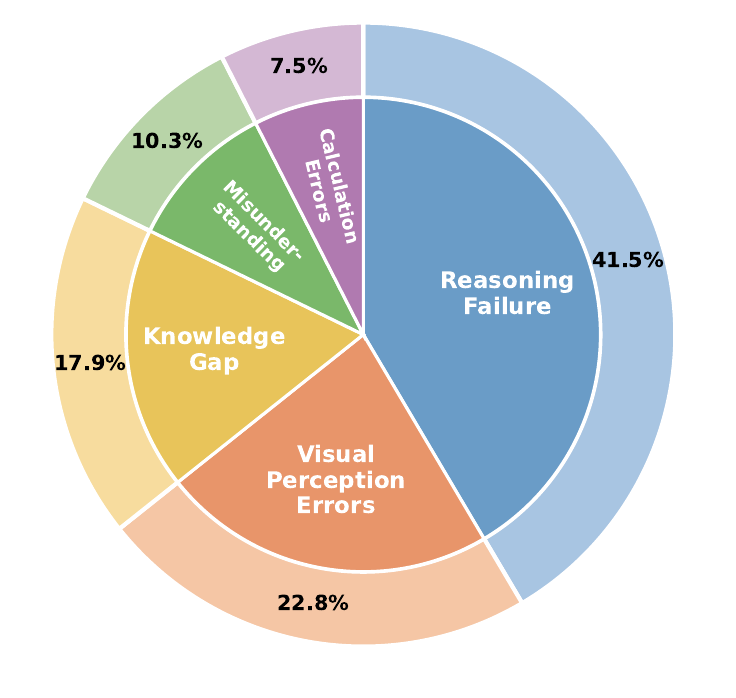}
    \caption{Error type distribution for Gemini~3.1~Pro on \benchname{}.}
    \label{fig:error-pie}
\end{figure}

To understand where the strongest model still fails, we classify errors made by Gemini-3.1-Pro on \benchname{} into five categories (Figure~\ref{fig:error-pie}).
\emph{Reasoning failure} is the dominant error type (41.5\%), where the model perceives visual content correctly but produces flawed logical inference, consistent with Visual Logic being the hardest dimension (\S\ref{sec:main-results}).
The remaining errors divide among \emph{visual perception} (22.8\%), primarily misidentifying symbols or fine-grained visual details; \emph{knowledge gaps} (17.9\%), mainly in Expert Knowledge and Medical dimensions; \emph{misunderstanding} (10.3\%); and \emph{calculation errors} (7.5\%). These results suggest that improving logical reasoning remains the most impactful direction.

\subsection{Effect of Reasoning Effort}
\label{sec:reasoning-effort}

\begin{table}[t]
    \centering
    \small
    \begin{tabular}{lccc}
    \toprule
    Dimension & Low & Medium & High \\
    \midrule
    STEM Reasoning      & 38.9 & 46.0 & 49.1 \\
    Expert Knowledge    & 53.9 & 57.9 & 60.1 \\
    Visual Logic        & 27.2 & 28.0 & 31.2 \\
    Visual Perception   & 59.2 & 60.0 & 60.7 \\
    Diagram \& OCR      & 64.3 & 66.0 & 66.2 \\
    Spatial             & 64.6 & 66.0 & 66.6 \\
    Medical             & 42.5 & 44.2 & 43.7 \\
    \midrule
    Macro Avg           & 50.1 & 52.6 & 53.9 \\
    \bottomrule
    \end{tabular}
    \caption{Cross-model average score by reasoning effort level across seven capability dimensions (5 models).}
    \vspace{-8pt}
    \label{tab:reasoning-effort}
\end{table}

We evaluate GPT-5, GPT-5-Mini, Gemini-3.1-Pro, Gemini-3.1-Flash-Lite, and Claude-Sonnet-4.6, at three effort levels (\textsc{Low}, \textsc{Medium}, \textsc{High}) following the protocol in \S\ref{sec:exp-setup}.

As shown in Table~\ref{tab:reasoning-effort}, increasing effort yields a Macro gain of $+$7.6\% from \textsc{Low} to \textsc{High}, but the benefit is concentrated in reasoning-intensive dimensions: STEM gains the most ($+$26.2\% relative), followed by Visual Logic ($+$14.7\%) and Expert Knowledge ($+$11.5\%).
In contrast, perception-oriented dimensions (Visual Perception, Diagram \& OCR, Spatial) remain nearly flat ($\leq$3.1\% gain), and Medical even declines slightly at \textsc{High}.
These results suggest that current models' perception performance is bounded by visual encoding quality rather than reasoning depth.

\section{Conclusion}

We present \benchname{}, a curated vision-language benchmark of 7,262 items across seven capability dimensions, distilled from 18 existing benchmarks through a systematic audit that identifies three pervasive quality issues: weak visual dependency, item saturation, and pseudo-hard questions. We address these issues with a reusable pipeline integrating text-only ablation, cross-model saturation detection, anomaly adjudication, and expert review. Evaluation on 27 LVLMs confirms that \benchname{} preserves model rankings while substantially improving discriminability. We release \benchname{} and the curation pipeline to support higher-quality multimodal evaluation.

\section*{Limitations}

\paragraph{Benchmark and Language Coverage.}
\benchname{} is constructed from 18 predominantly English benchmarks using closed-form formats (multiple-choice and fill-in-the-blank). This design choice ensures fully reproducible, scorer-consistent quality control, which is a prerequisite for our filtering pipeline. Extending the pipeline to open-ended generation tasks or multilingual settings would require adapting the scoring and filtering protocols accordingly, which we leave to future work.

\paragraph{Model Panel Dependency.}
Both saturation filtering and anomaly detection are conditioned on the 12-model panel used in this study. As future models improve, items currently retained may become saturated, and items currently flagged as anomalous may be resolved by more capable models. We mitigate this by covering models from five providers across three capability tiers, but the curated suite reflects the capability frontier at the time of construction. Our pipeline makes periodic re-curation with updated model panels straightforward.

\paragraph{Subjectivity in Anomaly Review.}
The anomaly detection stage combines rule-based candidate recall, multi-model adjudication by five LVLMs, and human expert review. While majority voting across independent judges reduces individual bias, the process retains inherent subjectivity: LLM judges may share systematic blind spots on certain domains, and human reviewers may disagree on genuinely ambiguous borderline cases. As a result, a different expert panel could reach different conclusions on a subset of items.

\section*{Ethical Considerations}

\paragraph{Data Provenance.}
All evaluation items in \benchname{} are sourced exclusively from previously published and publicly available benchmarks. We have reviewed the licensing terms of each source benchmark and confirmed that our use is consistent with their original intended purposes.

\paragraph{Sensitive Content.}
\benchname{} retains items from two medical benchmarks, SLAKE~\cite{SLAKE} and MedXpertQA~\cite{MedXpertQA}. These items are included solely for evaluating model capabilities in the medical domain and do not constitute medical advice. Both source datasets were de-identified and released by their original creators in compliance with applicable privacy regulations; we do not introduce any additional personally identifiable information.

\paragraph{Potential Biases.}
Our quality control pipeline relies on large models for text-only ablation, saturation detection, and multi-model adjudication. Although we mitigate single-model bias by aggregating judgments from multiple models across different providers, the filtering process may still reflect shared biases of current models, potentially favoring certain question types or penalizing others. Similarly, the human expert review stage, while serving as a final safeguard, may introduce annotator-specific biases. We encourage future work to investigate and quantify these effects.

\paragraph{Intended Use.}
\benchname{} is designed for academic evaluation of large vision-language models. Benchmark scores reflect model performance on curated evaluation items under controlled settings and should not be interpreted as indicators of real-world deployment readiness. Rankings may vary under different evaluation protocols, prompting strategies, or domain-specific configurations.

\paragraph{Computational Cost.}
Constructing \benchname{} requires substantial API calls and GPU computation for multi-model sampling and adjudication. However, we note that the resulting curated suite substantially reduces ongoing evaluation costs for future researchers: by retaining only high-quality, discriminative items, \benchname{} achieves comparable evaluation fidelity with significantly fewer items than evaluating on the full set of source benchmarks.

\bibliography{custom}

\end{document}